\documentclass[conference]{IEEEtran}
\IEEEoverridecommandlockouts

\usepackage[utf8]{inputenc}        
\usepackage{array}                 
\usepackage{multirow}              
\usepackage{csquotes}              

\usepackage[T1]{fontenc}
\usepackage[nocompress]{cite} 
\usepackage{microtype}
\usepackage{balance}
\usepackage{graphicx}
\usepackage{enumitem}
\usepackage{amssymb}

\usepackage{url}
\usepackage{xcolor}
\definecolor{dark-red}{rgb}{0.5,0.1,0.1}
\definecolor{dark-green}{rgb}{0.1,0.5,0.1}
\definecolor{dark-blue}{rgb}{0.1,0.1,0.8}
\usepackage[colorlinks,linkcolor=dark-red,citecolor=dark-blue,urlcolor=dark-blue]{hyperref}
\usepackage[nameinlink]{cleveref}

\usepackage{tikz} 
\newcommand\copyrighttext{%
  \scriptsize \textcopyright 2024 IEEE. Personal use of this material is permitted. Permission from IEEE must be obtained for all other uses, in any current or future media, including reprinting/republishing this material for advertising or promotional purposes, creating new collective works, for resale or redistribution to servers or lists, or reuse of any copyrighted component of this work in other works. Cite this article as follows: V. Švábenský, K. Tkáčik, A. Birdwell, R. Weiss, R. S. Baker, P. Čeleda, J. Vykopal, J. Mache, A. Chattopadhyay. \textit{Detecting Unsuccessful Students in Cybersecurity Exercises in Two Different Learning Environments}. In Proceedings of the 54th IEEE Frontiers in Education Conference (FIE '24). Washington, D.C., USA, 2024. DOI: \href{https://doi.org/10.1109/FIE61694.2024.10893135}{10.1109/FIE61694.2024.10893135}.}
\newcommand\copyrightnotice{%
\begin{tikzpicture}[remember picture,overlay]
\node[anchor=south,yshift=5pt] at (current page.south) {\fbox{\parbox{\dimexpr\textwidth-\fboxsep-\fboxrule\relax}{\copyrighttext}}};
\end{tikzpicture}%
}

\newcommand{\numkypodata}{244}
\newcommand{\numedurangedata}{69}
\newcommand{\numtotaldata}{313}

\newcolumntype{C}[1]{>{\centering\let\newline\\\arraybackslash\hspace{0pt}}m{#1}}

\newcommand{\kypo}{KYPO CRP}
\newcommand{\kypouni}{three European universities}
\newcommand{\edurange}{EDURange}
\newcommand{\edurangeuni}{Lewis \& Clark and The Evergreen State College}

\begin{document}

\title{Detecting Unsuccessful Students in Cybersecurity Exercises in Two Different Learning Environments
\copyrightnotice
\thanks{Funded by the European Union (grant number 101087529) and the National Science Foundation (grant numbers 2216492 and 2216485).}
}


\author{
    \IEEEauthorblockN{\textbf{Valdemar Švábenský}}
    \IEEEauthorblockA{University of Pennsylvania\\
    Pennsylvania, USA\\
    {\small valdemar.research@gmail.com}}
\and
    \IEEEauthorblockN{\textbf{Kristián Tkáčik}}
    \IEEEauthorblockA{Masaryk University\\
    Czech Republic\\
    {\small tkacikk@mail.muni.cz}}
\and
    \IEEEauthorblockN{\textbf{Aubrey Birdwell}}
    \IEEEauthorblockA{Georgia Institute of Technology\\
    Georgia, USA\\
    {\small aubrey.birdwell@gatech.edu}}
\and
    \IEEEauthorblockN{\textbf{Richard Weiss}}
    \IEEEauthorblockA{The Evergreen State College\\
    Washington, USA\\
    {\small weissr@evergreen.edu}}
\and
    \IEEEauthorblockN{\textbf{Ryan S. Baker}}
    \IEEEauthorblockA{University of Pennsylvania\\
    Pennsylvania, USA\\
    {\small ryanshaunbaker@gmail.com}}
\and
    \IEEEauthorblockN{\textbf{Pavel Čeleda}}
    \IEEEauthorblockA{Masaryk University\\
    Czech Republic\\
    {\small celeda@fi.muni.cz}}
\and
    \IEEEauthorblockN{\textbf{Jan Vykopal}}
    \IEEEauthorblockA{Masaryk University\\
    Czech Republic\\
    {\small vykopal@fi.muni.cz}}
\and
    \IEEEauthorblockN{\textbf{Jens Mache}}
    \IEEEauthorblockA{Lewis \& Clark College\\
    Oregon, USA\\
    {\small jmache@lclark.edu}}
\and
    \IEEEauthorblockN{\textbf{Ankur Chattopadhyay}}
    \IEEEauthorblockA{Northern Kentucky University\\
    Kentucky, USA\\
    {\small chattopada1@nku.edu}}
}

\maketitle

\begin{abstract}
This full paper in the research track evaluates the usage of data logged from cybersecurity exercises in order~to predict students who are potentially at risk of performing poorly. Hands-on exercises are essential for learning since they enable students to practice their skills. In cybersecurity, hands-on exercises are often complex and require knowledge of many topics. Therefore, students may miss solutions due to gaps in their knowledge and become frustrated, which impedes their learning. Targeted aid by the instructor helps, but since the instructor's time is limited, efficient ways to detect struggling students are needed. This paper develops automated tools to predict when a student is having difficulty. We formed a dataset with the actions of \numtotaldata\ students from two countries and two learning environments: \kypo\ and \edurange. These data are used in machine learning algorithms to predict the success of students in exercises deployed in these environments. After extracting features from the data, we trained and cross-validated eight classifiers for predicting the exercise outcome and evaluated their predictive power. The contribution of this paper is comparing two approaches to feature engineering, modeling, and classification performance on data from two learning environments. Using the features from either learning environment, we were able to detect and distinguish between successful and struggling students. A~decision tree classifier achieved the highest balanced accuracy and sensitivity with data from both learning environments. The results show that activity data from cybersecurity exercises are suitable for predicting student success. In a potential application, such models can aid instructors in detecting struggling students and providing targeted help. We publish data and code for building these models so that others can adopt or adapt them.
\end{abstract}


\section{Introduction}
\label{sec:intro}

As cyber threats become increasingly complex, organizations have big demand for cybersecurity experts~\cite{isc2022}. In order to train more experts, effective teaching methods such as hands-on exercises must be employed at universities and in professional learning contexts. However, cybersecurity exercises incorporate a wide range of topics, including operating systems, security vulnerabilities, and proficiency with command-line tools and programming languages. Due to the exercises' complexity, students often get stuck or frustrated~\cite{vinlove2021, chung2014}, which discourages them and hinders their learning.

Therefore, it is crucial for instructors to know when a student is at risk of not completing an exercise. This should be detected quickly -- the instructors' time is limited, so it should be invested in helping students who need it the most. Moreover, not all students will let instructors know when they need support, or even deny needing it, so the instructors may be unaware that a specific student is struggling. In these cases, automated tools help focus instructors' attention~\cite{Koutcheme2022methodological}. This enables instructors to target teaching interventions, such as hints, with the goal of positively impacting student learning.

\subsection{Goals and Scope of This Paper}

Our goal is to extract information from student actions in cybersecurity exercises, in order to predict student success or a potential risk of performing poorly. We evaluate tools that highlight students who may need help with an exercise task.

In our context, we broadly define successful students as being able to complete a certain amount of the exercise tasks; \Cref{subsec:methods-labeling} explains the specifics. Our work focuses on the education of cybersecurity students at the university level or beyond, though it could also be adapted to K-12 contexts.

This paper poses two research questions (RQ):
\begin{enumerate}
    \item How well do different machine learning classifiers predict (un)successful students in cybersecurity exercises?
    \item Are the best classifiers in one context also the best in another context, when trained using the same methods with a second student population in different exercises?
\end{enumerate}

\subsection{Contributions to Research and Practice}

First, we collected an original dataset from a total of \numtotaldata\ students in two learning environments. Second, we automatically extracted two feature sets from these data. Third, we used these features to train and evaluate eight types of binary classification models for predicting student success in the exercises. Feature selection and hyperparameter tuning were conducted for each model using nested cross-validation. To support reproducibility and replicability of the research, as well as recent calls for open educational data~\cite{Kiesler2023}, the data and code for this paper are publicly available (see \Cref{subsec:materials}).

Our choice of research methods is a response to the calls made in an extensive literature review on student performance prediction~\cite{Hellas2018}. Its authors urged the community to:
\begin{itemize}
    \item \textit{Use data from a second population at another institution}, since only 28 of 357 (7.8\%) articles assessed the prediction in more contexts. In a review aimed at CS1~\cite{Quille2019cs1}, only 2 of 47 (4.3\%) studies were conducted in more institutions.
    \item \textit{Use data from multiple courses and semesters} to avoid single-course single-semester case studies, which dominate the current literature but have limited generalizability. In~\cite{Quille2019cs1}, only 7 of 47 (14.9\%) studies were conducted over the period of more than one year.
    \item \textit{Publish the data collection instruments and the dataset itself}. Almost no datasets from previous studies were made public, and only 91 of 357 (25.5\%) articles included the data collection instruments~\cite{Hellas2018}.
\end{itemize}

This study adopts a multi-national, multi-institutional approach that overcomes many shortcomings of previous computing education papers~\cite{handbook-CER1}. It involves two distinct learning platforms, two sets of exercises, and student populations from two continents. This setup helps us to make progress towards understanding which techniques work in general for the problem of detecting unsuccessful students. 



\section{Related Work in Predictive Models}
\label{sec:related-work}

This section provides an overview of studies that analyzed data from educational contexts to predict student performance.
We discuss the state of the art in cybersecurity education (\Cref{subsec:related-work-cyber}) and in other computing education domains (\Cref{subsec:related-work-other}). Then, \Cref{subsec:related-work-gaps} summarizes the literature gaps that our work addresses.
The model evaluation metrics discussed here are defined in detail in \Cref{subsubsec:model-metrics}.

\subsection{Hands-on Cybersecurity Education}
\label{subsec:related-work-cyber}

Cybersecurity education literature employs student data to achieve different goals~\cite{my-2022-slr, Macak2022}, such as clustering students into high- and low-performing based on students' strategies for completing exercises~\cite{Koutcheme2022exploring}. However, the problem of predicting student performance is not covered much in previous research.

Vinlove et al.~\cite{vinlove2021} detected at-risk students in a command-line-based security exercise. They collected logs from 25 students and extracted three features: (1) average number of commands per task, (2) longest repeat of a command, and (3) edit distance between the command and \textquote{known-good} commands for the exercise. A support vector machine (SVM) reached 80\% accuracy in classifying exercise completion. However, this work used a rather small dataset and only one model.

Deng et al.~\cite{deng2018} predicted course performance of 103 students based on their activity (e.g., executed commands or mouse clicks) in a learning environment. The behavioral data, combined with real-time assessments, were used to train a naive Bayes classifier to predict students' grade category. At-risk students (the worst grades) were predicted with 90.9\% accuracy. Similarly to Vinlove et al.~\cite{vinlove2021}, this study evaluated only one classifier and reported only the accuracy metric.

Silva et al.~\cite{silva2014} searched for factors impacting the performance of 11 professionals in cybersecurity exercises. Shorter time gaps between participants' answer submissions correlated with submitting incorrect answers, which led to higher task abandonment. Also, participants who often switched between tools submitted more incorrect answers. However, unlike the other two papers, this work explored only statistical models.

\subsection{Other Areas of Computing Education}
\label{subsec:related-work-other}

Beyond the cybersecurity context, research on student success prediction is considerable. Hellas et al.~\cite{Hellas2018} reviewed 357 articles published by 2018 on predicting performance in computing courses, noting that the best papers \textquote{utilized data from multiple contexts and compared multiple methods}. The shortcomings of the reviewed papers were: lack of clearly defined research questions, single student population, limited discussion of validity, and little sharing of research data. Our work aims to address these shortcomings.

Koutcheme et al.~\cite{Koutcheme2022methodological} highlighted methodological issues in predicting unsuccessful students as they progress through a week-by-week course. The research uses student data to predict early in the semester whether a certain student is at risk of becoming inactive in the following week(s). The authors argue that if such predictive models are built using data of all students, including those who have already dropped out (an \textit{including} approach), the model performance is inflated because the inactive students stop generating new data, which simplifies the model decision-making (i.e., no activity $\rightarrow$ likely unsuccessful). This way, the model has access to \textquote{future} information it would otherwise not have in a realistic situation during the semester. Instead, they argue for an \textit{excluding} approach: to predict performance in week $n + 1$ such that only the data of students who are still active in week $n$ are included.

The remainder of our literature review covers papers after 2018 to complement Hellas et al.~\cite{Hellas2018}. We divide the publications based on whether they focus on individual exercises or whole courses. The former scope is closer to our work, but it has been rarely explored -- the latter scope dominates the literature.

\subsubsection{Prediction of Student Success in Exercises}

Arakawa et al.~\cite{arakawa2022} argued that predicting student performance at the course level provides insufficient insight into specific student struggles. They also suggested that features specific to the course or the student demographics may hinder replicating and automating the prediction. Instead, they identified struggling students at the level of programming assignments, using features such as the number of added code lines and the ratio of the passed tests. The study compared four models trained on data of 312 students; the best-performing one was a long short-term memory (LSTM) network with an AUC of 92.2\%.

Hicks et al.~\cite{hicks2022} explored whether the student success on \textit{in-class} coding exercises can predict success on weekly \textit{lab} coding exercises. The data for training four predictive models reflected the performance of 300 students in previous in-class exercises. The data captured two aspects: \textit{recency} (time between the in-class exercise and the lab exercise) and \textit{relevancy} of the in-class exercise to the lab exercise. The performance in recent relevant in-class exercises was the best predictor of success, with the corresponding Random Forest model reaching 84\% accuracy and 77\% precision and recall.

\subsubsection{Prediction of Student Success in Courses}

Koutcheme et al.~\cite{Koutcheme2022methodological} investigated student performance in three computing courses to compare the including and excluding approach (see \Cref{subsec:related-work-other}). Using the data of at least 1657 students, they extracted six types of features to train four models. The performance was better for the including approach. However, the results were inconclusive as to whether certain features are better for either of the two approaches.

Quille et al.~\cite{Quille2019cs1} presented 13 years of evolution of a model that employed a multi-institutional dataset for early prediction of student success in introductory programming. Using the data of 692 students, the Naive Bayes classifier (the best out of six models) achieved prediction accuracy of 71\%, sensitivity of 75\%, and specificity of 66\%. In a follow-up work~\cite{Quille2022}, they used a dataset of 472 students that was also multi-national. The three measures of the Naive Bayes classifier improved to 78\%, 79\%, and 78\%. In addition, a decision tree classifier reached 89\%, 90\%, and 89\%, respectively. 

Gao et al.~\cite{gao2021} mined patterns in process data of 106 students from a programming environment. The patterns were used to train an AdaBoost classifier that predicted student course outcomes with 79\% accuracy.

Leinonen et al.~\cite{leinonen2022} compared two time-on-task metrics: \textit{coarse-grained} (first keystroke to first exercise submission) and \textit{fine-grained} (sum of delays between all keystrokes until the first submission). The latter was a stronger predictor of performance. Using data of 132 students, the best model out of three was a Random Forest classifier (97\% AUC).

Gordon et al.~\cite{Gordon2023} analyzed data from a learning system for programming. They studied correlations between metrics (such as completion of reading assignments) and student exam performance. A decision tree predicted students at risk of failing the exam with 82\% sensitivity and 89\% specificity. 

Edwards et al.~\cite{Edwards2020} used students' keystroke data from two different programming courses at two institutions. The Python course at a US university had 265 students, and the Java course at a Finnish university had 303 students. Two Random Forest models (one for each course) were trained using 10-fold cross-validation in 10 runs with different random seeds. The average accuracy of predicting outcomes was 62\% in the Python course and 68\% in the Java course. Reducing the dataset only to students who attended the course exam improved the latter model (72\%), but did not change the former model.

Higueras et al.~\cite{higueras2018} predicted performance in a computer science course based on 86 students' interaction with a version control system. The prediction features included the number of commits or code additions/deletions. Naive Bayes and Random Forest were the best-performing classifiers trained on these features, scoring slightly above 80\% in accuracy and $F_1$ score.

\subsection{Literature Gaps and Novelty of This Paper}
\label{subsec:related-work-gaps}


The novel contributions of our study and unexplored areas compared to the prior work are discussed below.

\subsubsection{Focus on Cybersecurity}

Liao et al.~\cite{Liao2019robust} point out that computing education research has focused on introductory programming, but much less is known about students in follow-up courses. We identified few studies about cybersecurity; 10 out of 13 reviewed papers examine programming education. Although predicting performance in cybersecurity might not considerably differ from programming exercises, since many prediction features are universal, the literature is missing the investigation of whether results in cybersecurity would differ from those in other areas. Quoting Liao et al.~\cite{Liao2019robust}, there is \textquote{much to gain by studying courses that have not received as much research attention to this point}.

\subsubsection{Application of Multi-Contextual Data}

Hellas et al.~\cite{Hellas2018} and Liao et al.~\cite{Liao2019robust} encourage researchers in prediction modeling to perform studies across institutions, curricula, and semesters. However, multiple datasets from different contexts were not used in previous related research in cybersecurity, and rarely in other domains. Only 3/13 papers included data from more than one institution; 3/13 investigated more than one course, and 4/13 collected data over more than one semester. Our study goes beyond previous work by evaluating prediction models across two learning environments in two countries, throughout multiple courses and semesters in different schools.

\subsubsection{Prediction in Smaller Time Frames}

9/13 papers predict success in a course that lasts several weeks. Few papers look at the scope of a single exercise. However, the prediction at this granularity, as in our study, is arguably more difficult than throughout a course, since students generate less data.

\subsubsection{Comparison of Various Methods}

The papers compared limited number of models, 4 of them reporting the results of only one classifier, often with a limited feature set. We extract a large number and variety of features from two types of security exercise log data to train and evaluate eight models. 

\subsubsection{Sharing of Research Artifacts}

Hellas et al.~\cite{Hellas2018} implore researchers to share research data and code to support reproducibility and replicability. However, only 2 of the reviewed papers shared code, and only 1 paper shared the data (from 25 students only)~\cite{vinlove2021}. We make our datasets publicly available.

\section{Research Methods}
\label{sec:methods}

\begin{figure}[!ht]
\centering
\includegraphics[width=\linewidth]{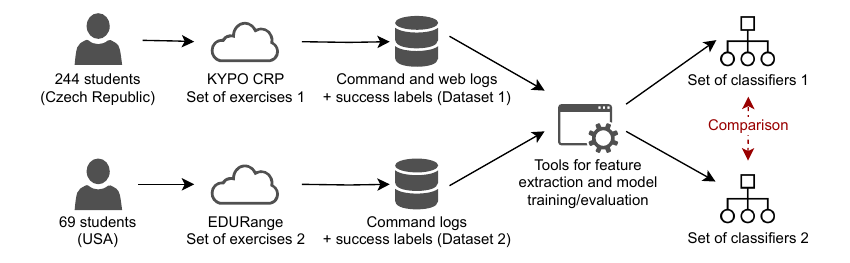}
\caption{Overview of the study design.}
\label{fig:overview}
\end{figure}

The term \textit{exercise} denotes a set of complex, multi-step \textit{tasks} in which the students practice cybersecurity skills. For example, the task can involve scanning open network ports of a computer system. As \Cref{fig:overview} shows, we studied several exercises with student populations in two learning environments: \kypo~\cite{vykopal2021} and \edurange~\cite{edurange_SIGCSE2015}. Our research quantitatively analyzes data from these two different contexts.

\subsection{Format and Content of the Cybersecurity Exercises}
\label{subsec:methods-environments}

In \kypo\ exercises, students breach vulnerable emulated hosts using a Kali Linux~\cite{kali-website} virtual machine (VM). For this study, we aggregated data from all 12 available exercises because all have the same underlying principles. The exercise instructions are presented via a web interface. In each exercise, the student must complete several linearly ordered tasks and obtain text strings representing the \textit{answers}. After submitting the correct answer to the web interface, the task is completed, and the student proceeds to the next task. If a student lacks the time or knowledge to complete a task, they can use the web interface to display the step-by-step \textit{solution}~\cite{kypo-documentation}.

\edurange\ includes attack and defense exercises. For this study, we chose the most thoroughly tested exercise, in which the students use Linux command-line tools in a VM to find files, change permissions, etc. The exercises are deployed via a web application and an SSH session, with instructions on the VM as text files. \edurange\ was designed independently from \kypo\ and differs in three aspects: (a) students can complete the tasks in any order, (b) step-by-step task solutions are not available, and (c) answers are not submitted and evaluated automatically in a web interface. Instead, \edurange\ automatically detects when a student has completed a task by comparing student data with known solutions.

In both platforms, the student has to complete complex tasks rather than multiple choice or short answer questions, so measuring progress is not straightforward. Like in programming, there can be multiple solutions, but there is no abstract syntax tree or unit tests to look at. Therefore, we use machine learning to discover patterns within these complex data of students.

\subsection{Data Collection in the Two Learning Platforms}

In \kypo, two data types are collected for each student: \textit{command logs} (all commands executed inside an interpreter such as Bash, Z shell, and Metasploit; with metadata such as timestamps)~\cite{collecting-shell-commands} and \textit{event logs} (interactions with the web interface of \kypo, such as submitting an answer or displaying a solution). The data collection instrument is open-source software~\cite{collecting-shell-commands}. An example dataset, including a detailed explanation of its format, is publicly available as well~\cite{svabensky2021}.

\edurange\ uses a single logging format that stores character stream data from a terminal. The command-line entries are automatically filtered and parsed, resulting in labeled log data. The first log entry of a student is triggered with the first submitted command. Each log entry consists of unique identifiers for the course, student, and task; a timestamp; and command input and output. Input is parsed into its command name (tool), options, and arguments. These pieces are compared, component-wise, to a set of known solutions, in order to match student behaviors to known actions.

\subsection{Data Collection for This Study}

\kypo\ accommodated \numkypodata\ students from May 2020 to May 2022. The participants were cybersecurity professionals, computer science students from \kypouni, and senior high school students. Some exercises were parts of the university courses on cybersecurity, where they constitute a mandatory lab session in which all students participate. No grade is given for the exercise, since it is only a practice session during the semester, and students are not penalized for mistakes. Other exercises were deployed during optional extracurricular activities such as educational events for popularizing cybersecurity -- hence the diverse population. Sessions of both types were either in-person or remote because some of them took place during the COVID-19 restrictions. Each exercise usually lasted 2--3 hours. In rare cases, some students finished the exercise at home within 24 hours of the start.

In \edurange, we collected data from \numedurangedata\ students during three undergraduate computer science courses offered at \edurangeuni\ between Fall 2020 and Fall 2022. Sometimes, the exercises were used as labs toward credit in these courses, and other times as optional workshops or extra credit assignments. Again, the learning modality (in-person or remote) varied. Most exercises were completed during a 2-hour lab session, but students were allowed to return to their work outside of the lab -- in some cases, even the next scheduled class session within 1--3 days.

During all exercises in both platforms, the students were allowed to use online resources, but were instructed to work individually. However, brief collaboration, such as when a student asked their neighbor a question, could not always be prevented. If a student needed help, and received it from another student, the exercise logs indicate a successfully solved task. As in most past studies, we cannot detect student cross-talk from the logs. However, since the goal is to recognize students who need timely help from the instructor, a student who can finish the task after asking for help from another student is ultimately classified correctly if they are classified as successful.

\subsection{Data Privacy and Research Ethics}

In both platforms, the participants received a written explanation that their anonymized exercise activity data may be used for educational research. Before starting the exercise, all participants whose data are included in this study gave informed consent, agreeing to this data collection. The collected data were manually checked to ensure they do not contain any personal information that could reveal a student’s identity. Likewise, the classification models that result from this work cannot leak any identifiable information about an individual student. Therefore, this research received a waiver from the institutional review boards of the involved universities.

\subsection{Data Cleaning and Filtering}

In addition to the data anonymization, multiple authors subjected the dataset to a thorough manual and automated inspection. We removed rare occurrences of unreasonable values in the data, such as the same command with the same timestamp logged multiple times, which were most likely caused by temporary network outages within the exercise platforms.
The final dataset used for the research includes:
\begin{itemize}
    \item For \numkypodata\ students in \kypo, we have 21,659 command logs and 8,690 event logs from the web interface.
    \item For \numedurangedata\ students in \edurange, we have 4,762 command logs (the platform does not use web event logs).
\end{itemize}

The combined sample includes data from \numtotaldata\ students over the period of more than two years across several semesters. This is well beyond the recommended minimal sample size for prediction studies in computing education: 96 students~\cite{Quille2019cs1}. We used this dataset for all the subsequent modeling steps.
While we acknowledge the potential risks of including the complete dataset for making predictions (see the discussion of Koutcheme et al.~\cite{Koutcheme2022methodological} in \Cref{subsec:related-work-other}), we argue that this decision makes sense in our context for three reasons.

First, Koutcheme et al.~\cite{Koutcheme2022methodological} operate in the time frame of weeks in a semester (and argue that this extrapolates to days in a week). When a student becomes inactive for several days, it is reasonable to assume that the student has most likely dropped out and exclude their data. However, our context is a single, one-time exercise session that lasts a few hours. When a student becomes inactive during this exercise (i.e., no action is logged for a certain time), the log data alone do not provide ground truth for deciding whether this student dropped out or will continue later. For example, the student can be discussing something with the instructor, experiencing technical difficulties with the exercise platform, or simply taking a small break.

Second, in our context, any criterion for rejecting student data as inactive would be arbitrary and hard to justify from the data (see Kovanović et al.~\cite{Kovanovic2016} discuss the challenges of choosing cut-offs for inactivity within interaction data).

Third, since our goal is to classify whether a student will finish the exercise, if we eliminated the data of students who stopped working, our training set would include (almost) exclusively successful students. So, the model would have (almost) no data based on which to learn to predict failure. Partial data of students who dropped out of the exercise are valuable, as they capture the behavior of unsuccessful students.

Ultimately, inactive students are a part of the learner population and naturally occur in teaching contexts. Thus, continuing to use such data better reflects the model's potential practical use. However, to provide an empirical comparison, we will also present results on a subset (half) of the dataset.

\subsection{Definition of Class Labels and Data Labeling}
\label{subsec:methods-labeling}

Our goal is to predict if a student needs help. The exercise outcome is represented with a binary class variable (label). This label is set to 1, which is a positive class that we want to detect, if a student was not successful (i.e., potentially at risk); and 0, the negative class, indicating the student succeeded. \Cref{tab:data_distr} shows the labels' distribution in the dataset.

\begin{table}[!ht]
\begin{center}
\caption{The distribution of positive and negative training labels in the datasets from the two learning environments.}
\label{tab:data_distr}
\def\arraystretch{1.25}
\begin{tabular}{lrrr}
              & Unsuccessful (1) & Successful (0) & Total \\ \hline
    \kypo     & 63 (26\%)      & 181 (74\%)     & \numkypodata \\
    \edurange & 14 (20\%)      &  55 (80\%)     & \numedurangedata \\
\end{tabular}
\end{center}
\end{table}

Exercise \textit{success} is defined as at least 50\% completion. In \kypo, this means that the student (a) did not display the solutions for more than 50\% of the tasks in the exercise and (b) submitted the correct answers for all tasks (whether discovered by the student or offered by the solution). In \edurange, this is simply finishing at least half of the tasks (rounding down when the exercise had an odd number of tasks).

While this cut-off is low, it was chosen because it represents a minimal completion of the exercises, and we want to focus on identifying students who need help the most, i.e., those who are unable to reach even 50\%. This is consistent with the goal of the exercises, which is to provide a learning experience to undergraduates (often cybersecurity beginners), not to evaluate student performance for a letter grade.

Although this choice makes the dataset imbalanced towards successful students, this reflects the settings in which both platforms were employed. The 50\% threshold is the lowest passing grade in most courses at the university where \kypo\ is used. In addition, the second half of the tasks are highly challenging compared to the first half, so setting the threshold higher might lead to more predictions of failure than would be actionable for the instructor.

\subsection{Feature Extraction and Selection}

For \kypo\ data, we engineered 25 features in preliminary work~\cite{Tkacik2022thesis}. The features were derived only from the exercise problem-solving. Like Edwards et al.~\cite{Edwards2020}, we did not use (or even collect) student personal information.
Next, all features were unitized (before model training to avoid data leakage). Then, we applied automated feature selection using L1-regularized linear models~\cite{scikit-feature-selection}, which pruned the feature set before the training phase of each model (see \Cref{subsubsec:crossval}). 
\Cref{tab:features_kypo} lists the 25 selected features and their descriptive statistics.
Most of the features are analogous to those identified in the review by Hellas et al.~\cite{Hellas2018} (e.g., time on task, number of attempts, and correctness), which are commonly used in the literature and also generalize outside the cybersecurity context.

\begin{table*}[t]
\scriptsize

\begin{center}
\caption{All 25 features used for building models from \kypo\ data. All values are computed from per-student data, $n = $ \numkypodata. 
}
\label{tab:features_kypo}
\def\arraystretch{1.25}
\begin{tabular}{crlrrrrr}
 & \textbf{\#} & \textbf{Feature description [unit or permissible range]} & \textbf{Min} & \textbf{Max} & \textbf{Med} & \textbf{Avg} & \textbf{Std} \\ \hline

\multirow{6}{*}{\rotatebox[origin=c]{90}{\textbf{Command usage}}} 
 & 1 & Avg number of commands executed per task [$\in \mathbb{R}_0^+$] & 0.8 & 71.6 & 12.7 & 15.4 & 11.4 \\
 & 2 & Min number of commands executed per task [$\in \mathbb{N}_0$]   & 0 &  10 &  0 &  1 &  2 \\
 & 3 & Max number of commands executed per task [$\in \mathbb{N}_0$]   & 2 & 211 & 35 & 46 & 37 \\
 & 4 & Avg number of commands executed per minute [$\in \mathbb{R}_0^+$] & $9e^{-6}$ & $2e^{-3}$ & $3e^{-4}$ & $3e^{-4}$ & $2e^{-4}$ \\
 & 5 & Length of longest sequence of repeated commands [$\in \mathbb{N}_0$] & 1 & 16 & 2 & 3 & 2 \\
 & 6 & Avg time gap between two command executions [m:ss] & 0:10 & 29:41 & 1:00 & 1:23 & 2:03 \\ \hline

\multirow{6}{*}{\rotatebox[origin=c]{90}{\textbf{Tool usage}}} 
 &  7 & Avg number of unique tools used per task [$\in \mathbb{R}_0^+$] & 0.6 & 19.6 & 4.3 & 5.0 & 3.1 \\
 &  8 & Min number of unique tools used per task [$\in \mathbb{N}_0$]   & 0 & 5 & 0 & 1 & 1 \\
 &  9 & Max number of unique tools used per task [$\in \mathbb{N}_0$]   & 1 & 66 & 11 & 13 & 10 \\
 & 10 & Avg number of unique tools used per minute [$\in \mathbb{R}_0^+$] & $7e^{-6}$ & $4e^{-4}$ & $7e^{-5}$ & $8e^{-5}$ & $5e^{-5}$ \\
 & 11 & Length of longest sequence of repeated tools [$\in \mathbb{N}_0$] & 2 & 37 & 6 & 7 & 5 \\
 & 12 & Avg time gap between using two different tools [m:ss]  & 0:11 & 59:53 & 1:29 & 2:21 & 4:24 \\ \hline

\multirow{4}{*}{\rotatebox[origin=c]{90}{\textbf{Solutions}}} 
 & 13 & Number of solutions displayed per task [0/1]                 & 0 & 1 & 0.17 & 0.25 & 0.28 \\
 & 14 & Solution displayed to a task in the first half of the exercise [0/1] & 0 & 1 & 0 & 0.37 & 0.48 \\
 & 15 & Solution displayed for two consecutive tasks [0/1]           & 0 & 1 & 0 & 0.28 & 0.45 \\
 & 16 & Min time from task start to solution displayed [m:ss] & 0:02 & 68:26 & 30:40 & 37:00 & 29:33 \\ \hline

\multirow{6}{*}{\rotatebox[origin=c]{90}{\textbf{Answers}}} 
 & 17 & Avg number of wrong answers per task [$\in \mathbb{R}_0^+$] & 0 & 11.3 & 0.5 & 0.7 & 1.1 \\
 & 18 & Max number of wrong answers without action performed [$\in \mathbb{N}_0$] & 0 & 23 & 1 & 2 & 2 \\
 & 19 & Avg time gap between two consecutive answers [m:ss]          & 0:37 & 248:08 & 8:22 & 11:12 & 16:26 \\
 & 20 & Avg time from task start to the first answer in that task [m:ss] & 3:34 & 199:15 & 14:01 & 16:23 & 14:05 \\
 & 21 & Min time from task start to the first answer in that task [m:ss] & 0:05 & 49:40 & 3:04 & 4:16 & 4:56 \\
 & 22 & Max time from task start to the first answer in that task [m:ss] & 7:33 & 824:58 & 31:44 & 60:53 & 53:29 \\ \hline

\multirow{3}{*}{\rotatebox[origin=c]{90}{\textbf{Duration}}} 
 & 23 & Avg time to successfully complete a task [m:ss] & 3:34 & 239:01 & 14:57 & 17:58 & 21:24 \\
 & 24 & Min time to successfully complete a task [m:ss] & 0:07 &  38:35 &  3:50 &  5:21 &  5:36 \\
 & 25 & Max time to successfully complete a task [m:ss] & 7:33 & 824:58 & 31:27 & 40:28 & 73:31 \\
 
\end{tabular}
\end{center}

\vspace*{-4mm}
\end{table*}

\edurange\ followed the same process, engineering 15 features in \Cref{tab:features_edurange}. The difference reflects the absence of web interface log data. On the other hand, \edurange\ produced features about command execution failure from the command-line output stream that were not available in \kypo. 

\begin{table*}[t]
\scriptsize

\begin{center}
\caption{All 15 features used for building models from \edurange\ data. The value is again always per student, $n = $ \numedurangedata. 
}
\label{tab:features_edurange}
\def\arraystretch{1.25}
\begin{tabular}{crlrrrrr}
 & \textbf{\#} & \textbf{Feature description [unit or range]} & \textbf{Min} & \textbf{Max} & \textbf{Med} & \textbf{Avg} & \textbf{Std} \\ \hline

\multirow{9}{*}{\rotatebox[origin=c]{90}{\textbf{Command usage}}} 
 & 1 & Avg number of commands executed per task [$\in \mathbb{R}_0^+$]   & 1.8 & 28.3 & 6.3 & 6.3 & 4.3 \\
 & 2 & Min number of commands executed per task [$\in \mathbb{N}_0$]     & 1 & 4 & 1 & 1 & 1 \\
 & 3 & Max number of commands executed per task [$\in \mathbb{N}_0$]     & 3 & 98 & 23 & 24 & 60 \\
 & 4 & Avg number of commands executed per minute [$\in \mathbb{R}_0^+$] & $3e^{-5}$ & 0.10 & 0.03 & 0.03 & 0.02 \\
 & 5 & Length of longest sequence of repeated commands [$\in \mathbb{N}_0$] & 1 & 85 & 2 & 5 & 13 \\
 & 6 & Avg time gap between two command executions [hh:mm:ss]          & 0:10 & 10:48:03 & 0:31 & 10:33 & 1:18:01 \\
 & 7 & Avg number of unique commands executed per task [$\in \mathbb{R}_0^+$]   & 1.7 & 4.6 & 2.9 & 2.9 & 0.6 \\
 & 8 & Min number of unique commands executed per task [$\in \mathbb{N}_0$]     & 1 & 2 & 1 & 1 & 1 \\
 & 9 & Max number of unique commands executed per task [$\in \mathbb{N}_0$]     & 2 & 12 & 7 & 7 & 2 \\ \hline

\multirow{2}{*}{\rotatebox[origin=c]{90}{\textbf{Errors}}} 
 & 10 & Avg number of errors per task [$\in \mathbb{R}_0$] & 0 & 3.4 & 0.4 & 0.6 & 0.6 \\
 & 11 & Max number of errors per task [$\in \mathbb{N}_0$] & 0 & 23 & 3 & 3 & 3 \\ \hline

\multirow{4}{*}{\rotatebox[origin=c]{90}{\textbf{Duration}}} 
 & 12 & Avg time to successfully complete a task [hh:mm:ss] & 2:16 & 50:24:55 & 9:52 & 56:37 & 6:02:39 \\
 & 13 & Min time to successfully complete a task [hh:mm:ss] & 0:00 & 5:23 & 0:00 & 0:27 & 1:00 \\
 & 14 & Max time to successfully complete a task [hh:mm:ss] & 3:49 & 75:36:18 & 24:06 & 1:36:10 & 9:03:15 \\
 & 15 & Avg time gap between two task completions [hh:mm:ss] & 0:52 & 25:12:06 & 2:40 & 25:15 & 3:01:40 \\
 
\end{tabular}
\end{center}

\vspace*{-5mm}
\end{table*}

\subsection{Model Training and Evaluation}

Datasets from both platforms were used for modeling separately, since the feature sets differ. However, we use the same analysis framework: the code for processing the datasets and training the models is the same, with slight adjustments to account for each platform’s specifics. Cross-platform research in learning analytics almost inherently has this property, unless features that do not perfectly map are discarded.

We systematically compared the performance of eight classifiers: \textit{logistic regression}, \textit{naive Bayes}, \textit{support vector machines} (with linear or RBF kernel), \textit{K-nearest neighbors}, \textit{decision tree} (CART), \textit{Random Forest}, and \textit{XGBoost}. We selected these standard models because, compared to deep learning models, simpler classifiers are more interpretable~\cite{Rudin2019stop}, which is suitable for educational purposes, and usually require less data. All implementations come from the Python library scikit-learn~\cite{scikit-learn}, only XGBoost has a separate package~\cite{xgboost}.

\subsubsection{Cross-validation}
\label{subsubsec:crossval}

We used nested student-level cross-validation~\cite{mlm-nested-cv, scikit-nested-cv} (rather than allocating a holdout test set). This method was chosen since \edurange\ dataset was smaller, and we wanted to use the same training process for both.

The \textit{inner} cross-validation loop is used for feature selection and hyperparameter tuning. Both procedures are performed on the training set, split into training and validation folds. The best model is selected by the \textit{outer} loop automatically among all models from the inner loop, by evaluating them on the test data (not used for feature selection or tuning for that fold).

Both loops use \textit{stratified} \textit{k}-fold cross-validation to account for the label imbalance.
The value for \textit{k} is commonly set to 10 for the outer loop~\cite{Quille2019cs1, Liao2019robust, Edwards2020}, which we chose as well, and a smaller number for the inner loop~\cite{mlm-nested-cv} (we chose 5).

Since the test set split is used only for model evaluation, not feature selection or hyperparameter tuning, and data from one student are not divided between the training and test set split (i.e., we use student-level cross-validation), there is no information leakage.

\subsubsection{Model Evaluation Metrics}
\label{subsubsec:model-metrics}

The metrics were chosen to maximize practical utility. We use \textit{sensitivity} to quantify the classifier's ability to identify struggling students, which is crucial so that the instructor can provide assistance. However, since the instructor's time is limited, it is also vital that students who do not need assistance are not reported as struggling. For this reason, we also consider \textit{specificity} -- the ability to predict successful students. These two metrics were also used in related publications~\cite{Quille2019cs1, Liao2019robust}. Quille et al.~\cite{Quille2019cs1} argue that these metrics are important, but rarely reported in related work (only in 4 out of 47 papers they reviewed).

When comparing models with a single metric (e.g., for hyperparameter tuning), we use \textit{balanced accuracy}~\cite{THOLKE2023} (the average of the sensitivity and specificity of the given model).
Finally, we compute the Area Under the Receiver Operating Characteristic Curve (\textit{AUC}), which is independent of decision thresholds and unaffected by data imbalance~\cite{Jeni2013}.

\subsubsection{Naive Baseline Models}

For comparison, we consider two trivial models. One is \textit{majority classification} -- each student is classified to the most prevalent class (0 in our case).
The other is \textit{random classification} -- each student is classified randomly.


\section{Results and Discussion}
\label{sec:results}

\Cref{tab:classif_comp1} reports the model performance for \kypo~and \Cref{tab:classif_comp2} for \edurange.
The values are macro-averages across the scores of the 10 models trained in the outer cross-validation loop. All eight models performed much better than the baseline.

\begin{table}[t]

\begin{center}
\caption{Classifier performance using the \numkypodata\ data points for \kypo. The models are sorted by the balanced accuracy descending.}
\label{tab:classif_comp1}
\def\arraystretch{1.25}
\begin{tabular}{lllll}
    Classifier           & Sensitivity & Specificity & Bal-Acc & AUC \\ \hline
    Decision tree        & \textbf{0.869} & 0.900 & \textbf{0.884} & 0.921 \\
    XGBoost              & 0.788 & 0.944 & 0.866 & 0.929 \\
    SVM (linear kernel)  & 0.838 & 0.884 & 0.861 & 0.909 \\
    Random Forest        & 0.759 & \textbf{0.961} & 0.860 & \textbf{0.931} \\
    SVM (RBF kernel)     & 0.776 & 0.911 & 0.843 & 0.901 \\
    Nearest neighbors    & 0.714 & \textbf{0.961} & 0.837 & 0.907 \\
    Naive Bayes          & 0.747 & 0.927 & 0.837 & 0.901 \\
    Logistic regression  & 0.776 & 0.889 & 0.832 & 0.889 \\
\end{tabular}
\end{center}

\begin{center}
\caption{Classifier performance using the \numedurangedata\ data points for \edurange. The models are sorted based on \Cref{tab:classif_comp1} for easier comparison.}
\label{tab:classif_comp2}
\def\arraystretch{1.25}
\begin{tabular}{lllll}
    Classifier           & Sensitivity & Specificity & Bal-Acc & AUC \\
    \hline
    Decision tree        & \textbf{0.900} & 0.740 & \textbf{0.820} & 0.826 \\
    XGBoost              & 0.500 & 0.816 & 0.658 & 0.821 \\
    SVM (linear kernel)  & 0.750 & 0.813 & 0.781 & 0.843 \\
    Random forest        & 0.700 & \textbf{0.873} & 0.786 & \textbf{0.853} \\
    SVM (RBF kernel)     & 0.850 & 0.700 & 0.775 & 0.796 \\
    Nearest neighbors    & 0.050 & 0.866 & 0.458 & 0.725 \\
    Naive Bayes          & 0.450 & 0.793 & 0.621 & 0.600 \\
    Logistic regression  & 0.800 & 0.733 & 0.766 & 0.828 \\
\end{tabular}
\end{center}

\vspace*{-3mm}

\end{table}

\subsection{RQ1: Classifier Performance in \kypo}
\label{results:kypo}

The decision tree has the highest balanced accuracy (88.4\%) and sensitivity (86.9\%). The best specificity (96.1\%) was achieved by Random Forest, which also had the best AUC (93.1\%), and Nearest neighbors, which was less suitable due to having the lowest sensitivity (71.4\%).

The decision tree is also the least biased toward a particular class (it has the smallest absolute difference between sensitivity and specificity, 3.1\%). This is desirable since we want to maximize the ability to detect at-risk students while minimizing the number of students incorrectly identified as struggling.

The difference between the balanced accuracy across the models is small: only 5.2\% between the highest and lowest-scoring classifier. Similarly, the difference between the highest and lowest AUC is low: 4.2\%. This suggests that while some classifiers are more suitable for the given context, none are entirely unusable, which is a good sign.


Surprisingly, experimenting with a subset of the dataset (using 50\% of each log file to simulate the students being somewhere in the middle of the exercise) did not deteriorate the models substantially. The balanced accuracy dropped by less than 0.1, and AUC by less than 0.05. In one model (logistic regression), the two metrics even marginally improved.

\subsection{RQ2: Comparison With \edurange}
\label{results:edurange}

Our next question was to evaluate if and how the results change when the methods are applied to another context. Despite differences in the number of students and features, \edurange\ results are generally consistent with \kypo. High values for all metrics were achieved by the same classifiers. Again, decision tree reached the highest balanced accuracy (82\%) and sensitivity (90\%). It also had an AUC of 82.6\%: well above the baseline. However, its 16\% difference between sensitivity and specificity indicates some bias.

Random Forest performed the second best, achieving a balanced accuracy of 78.6\% and an AUC of 85.3\%. Both SVMs performed well, with a balanced accuracy of 78.5\% and 78.1\%. SVM with the linear kernel was favorable given the distribution of our dataset, since it had lower absolute difference between sensitivity and specificity (6.3\%) and higher AUC (84.3\%).




Since \kypo\ uses features derived from its web interface, the two systems favored different features for the most part, but \textit{commands per minute} and \textit{the average number of commands used to complete a task} were used in both contexts. For \edurange, those features were chosen by many models.

Using 50\% of the dataset again deteriorated the models only slightly (balanced accuracy by up to 0.13 and AUC by up to 0.09), and even improved several models.

Despite some differences, the \edurange\ features were largely equivalent to \kypo\ when possible. Overall, the feature sets are comparably rich. \edurange\ ran the modeling with fewer data, and its logs captured only command-line activity. The slightly worse performance indicates that having additional web interface data improves the predictive power.

\subsection{Limitations and Threats to Validity}


\subsubsection{Internal Validity}
\label{subsubsec:results-internal-validity}

Any threshold that separates successful and at-risk students (including our setting of 50\% completion) is an arbitrary choice that affects the results. However, there is no theoretical basis or a \textquote{gold standard} in the literature to determine exactly when to consider a student as struggling~\cite{Koutcheme2022methodological}.

Regardless of where the cut-off is set, students who are near the threshold may belong in either category (e.g., two similar students who achieve a score of 51\% and 49\% end up classified into separate categories). As a result, the class labels (\textquote{unsuccessful} and \textquote{successful}) should be treated with some caution. However, this is a limitation of any binary classification. For example, Castro-Wunsch et al.~\cite{Castro-Wunsch2017} also used a 50\% cut-off. Edwards et al.~\cite{Edwards2020} used a median split, predicting whether the student will be in the top or bottom half performance-wise. Liao et al.~\cite{Liao2019robust} defined a 40\% cut-off, remarking that it can be adjusted to \textquote{trade off the sensitivity and specificity of a given model}. Ultimately, the threshold will vary according to the context relevant for a particular course or exercise.


\subsubsection{External Validity}

Exercise sessions sometimes differed in aspects such as student demographics, instructor, or modality. However, these changes are natural in field research in computing education. As Liao et al.~\cite{Liao2019robust} argue, it is unrealistic to expect that all conditions will remain constant across all teaching sessions. Moreover, due to the different design of the two platforms (which were developed before this study was considered), the two feature sets were not the same, despite having a substantial overlap. On the positive side, these differences may enhance generalizability.

Results of computing education research might not always transfer to a different context~\cite{ihantola2015}. Our models were trained and evaluated only on the two presented datasets. Therefore, we cannot make reliable claims regarding the generalizability to other exercises or platforms. Nevertheless, if exercises in other platforms allow quantifying student success, our methods can be applied with minimal modifications, since the source code is available (see \Cref{subsec:materials}).



\subsection{Implications for Teaching Practice}

The classifiers can be trained for other exercise environments and then deployed to detect unsuccessful students. To illustrate, suppose the best model was deployed in \kypo. The model has a sensitivity of 0.869, meaning that of all at-risk students, it can correctly classify 86.9\% of them. Next, the model has a specificity of 0.900, so it can correctly classify 90\% of all students who are actually successful. Finally, given a successful and unsuccessful student, the model with the AUC of 0.921 can accurately distinguish them in 92.1\% of cases. 

As evidenced in the raw logs, students perform hundreds of exercise actions, which is far too much for the instructor to evaluate manually. Therefore, the detectors would help direct the instructor's attention to students who need advice. Even if some students are misclassified, the rate of false positives/negatives is manageable in hands-on cybersecurity courses, which tend to have dozens (not hundreds) of students. 

A learning environment providing this detection would allow instructors to interact with students on a one-to-one basis, even within a large exercise, in-person or online. It can also counteract implicit bias because it removes the need for students to request help and for instructors to choose whom to monitor. The classification algorithms have no explicit information about demographics, as recommended by recent literature~\cite{Baker2023}.

From the technical perspective, developers of cybersecurity learning environments can implement logging of data that produce the most significant features. These include the number of commands, errors made, and the timing of answer submissions, as well as the exercise metadata, such as requesting solutions.

\section{Conclusions}
\label{sec:conclusions}

Identifying students who are at risk of performing poorly is essential for providing targeted interventions. To the best of our knowledge, only a few studies explored student performance prediction in cybersecurity exercises, and none of these studies were conducted across two platforms. We attempted to bridge this gap by using student activity data from \kypo\ and \edurange\ platforms to determine student success. Specifically, we employed classification to assess how well the features extracted from the activity data predict exercise outcomes. Evaluating eight models for the two platforms demonstrated that predicting student success based on exercise data is a promising approach that generalizes across contexts.

\subsection{Open Research Challenges}


The goals of the prediction can be modified in various ways. For example, future work can aim to discover specific tasks in the exercise on which the student will struggle. Classifying whether a student will complete a particular task eliminates the need to set arbitrary thresholds for success. Alternatively, the goal can be reframed as ranking the students based on how likely they are to need help, determining the priority for the instructor. Ultimately, future work should evaluate the practical deployment of these methods in a classroom.

Another scope of future work is detecting at-risk students as early as possible. The earlier an accurate prediction is made, the more beneficial it is~\cite{Quille2019cs1, Koutcheme2022methodological}, since instructors can intervene to support students quickly. Others explored this problem within an entire semester, achieving promising results using week-by-week data~\cite{Liao2019robust, Koutcheme2022methodological}. However, we are not aware of publications in a security context within a smaller time frame of a short exercise. An open challenge is therefore employing meaningfully selected data only from a subset of the exercise.

\subsection{Publicly Available Supplementary Materials}
\label{subsec:materials}

We publish the datasets, as well as scripts for processing exercise logs, extracting features, and training classifiers. The code is documented and can be extended to extract additional features or evaluate more models. For \kypo, see \url{https://gitlab.fi.muni.cz/cybersec/papers/2024-FIE-unsuccessful-students}. For \edurange, see \url{https://github.com/aubreybirdwell/2024-FIE-unsuccessful-students}.

\section*{Acknowledgment}
The researchers from Masaryk University were supported by the European Union under Grant Agreement No. 101087529. Part of this paper is based upon work supported by the National Science Foundation under grant numbers 2216492 and 2216485.

\newpage        
\balance
\bibliographystyle{IEEEtran}
\bibliography{references}

\end{document}